\documentclass[letterpaper, 10pt, journal, twoside]{IEEEtran}

\IEEEoverridecommandlockouts 

\makeatletter

\let\proof\@undefined
\let\endproof\@undefined
\makeatother

\usepackage{hyperref}
\hypersetup{
  colorlinks,
  citecolor=black,
  filecolor=black,
  linkcolor=black,
  urlcolor=black,
  pdfauthor={},
  pdfsubject={},
  pdftitle={}
}
\usepackage{amsmath, bm} 
\usepackage{amssymb}
\usepackage{pdfpages}
\usepackage{subcaption}
\usepackage{siunitx}
\usepackage{xcolor}
\usepackage{url}
\usepackage[]{hyperref}
\usepackage{cite}
\usepackage{pifont}
\usepackage{multirow}
\usepackage[]{algorithm}
\usepackage[noend]{algpseudocode}

\usepackage{comment}

\usepackage{graphicx}
\usepackage{setspace}
\usepackage{epstopdf}
\usepackage{float}
\usepackage{multirow,tabularx,makecell}
\usepackage{siunitx}
\usepackage{cellspace}
\usepackage[normalem]{ulem}
\usepackage{pifont}

\usepackage{amsmath}
\usepackage{amssymb}
\usepackage{mathtools}

\newcommand\referencediag[4]{{}^{{#2}}_{{#4}}{{#1}}^{}_{{#3}}}   
\newcommand\referencediagt[5]{{}^{{#2}}_{{#4}}{{#1}}^{#5}_{{#3}}}   
\newcommand\reference[4]{{\referencediag{#1}{#2}{#3}{#4}}}  
\newcommand\referencet[5]{{\referencediagt{#1}{#2}{#3}{#4}{#5}}}  
\newcommand{\refframe}[1]{\left\{\mathcal#1\right\}}

\newcommand\transpose{{{T}}}

\newcommand\bZero{\boldsymbol{0}}

\newcommand\bSigma{{\mathbf{\Sigma}}}

\makeatletter
\newcommand{\generateAlphabet}[4]{%
  \def\@tempa{#1} 
  \count@=`#3
  \loop
  \begingroup\lccode`?=\count@
  \lowercase{\endgroup\@namedef{\@tempa ?}{#2{?}}}%
  \ifnum\count@<`#4
  \advance\count@\@ne
  \repeat
}

\generateAlphabet{v}{\mathbf}{A}{Z}
\generateAlphabet{v}{\mathbf}{a}{z}
\generateAlphabet{c}{\mathcal}{A}{Z}
\generateAlphabet{cc}{\mathcal}{a}{z}
\generateAlphabet{cs}{\scriptscriptstyle\mathcal}{A}{Z}
\generateAlphabet{fr}{\mathfrak}{A}{Z}
\generateAlphabet{fr}{\mathfrak}{a}{z}
\generateAlphabet{rm}{\mathrm}{A}{Z}
\generateAlphabet{rm}{\mathrm}{a}{z}

\newcolumntype{D}{>{\hfill}N{3}{2}<{\hfill}}

\usepackage{algorithm}
\usepackage{etoolbox}\AtBeginEnvironment{algorithmic}{\scriptsize}
\usepackage[noend]{algpseudocode}

\usepackage{wrapfig}
\usepackage[font={small}]{caption}
\usepackage{scalefnt}

\usepackage[export]{adjustbox}
\usepackage{environ}
\usepackage[para, bottom]{footmisc}
\usepackage{color}
\usepackage{url}

\newcommand{\remove}{}

\usepackage{amsmath}
\usepackage{amssymb,amsmath}
\usepackage{mathtools}
\usepackage{siunitx}
\usepackage[nameinlink, noabbrev, capitalize]{cleveref}
\usepackage{cite}

\usepackage{trimclip}

\usepackage{isotope}
\usepackage{listings}
\makeatletter
\def\lst@makecaption{%
  \def\@captype{table}%
  \@makecaption
}
\makeatother

\makeatletter
\def\BState{\State\hskip-\ALG@thistlm}
\makeatother

\definecolor{dark_green}{rgb}{0.0, 0.6, 0.0}

\newcommand{\rev}[1]{{\color{black} {#1}}}

\newcommand{\PUBLISHEDIN}{IEEE Robotics and Automation Letters}
\newcommand{\DOI}{10.1109/LRA.2025.3562786} 
\usepackage[placement=top,vshift=-7]{background}
\SetBgScale{1.0}
\SetBgContents{\PUBLISHEDIN. PREPRINT VERSION - DO NOT DISTRIBUTE. \href{https://doi.org/\DOI}{DOI \DOI}}
\SetBgColor{black}
\SetBgAngle{0}
\SetBgOpacity{1.0}

\usepackage{tikz}
\usepackage{pgfplots}
\usepgfplotslibrary{groupplots}
\usetikzlibrary{backgrounds,arrows,automata,shapes,positioning,calc,through}
\pgfplotsset{compat=newest}
\pgfdeclarelayer{background}
\pgfdeclarelayer{foreground}
\pgfsetlayers{background,main,foreground}

\tikzset{
  state/.style={
    rectangle,
    draw=black, very thick,
    minimum height=1.0em,
    text centered,
  },
  legend_box/.style={
    rectangle,
    draw=black,
    text centered,
  },
  normalstate/.style={
    rectangle,
    draw=black, very thick,
    minimum height=2.9em,
    minimum width=6.25em,
    text centered,
  },
  finalstate/.style={
    rectangle,
    double=white,
    double distance=0.1em,
    inner sep=0.2em,
    draw=black, very thick,
    minimum height=2.90em,
    minimum width=6.25em,
    text centered,
  },
  initialstate/.style={
    rectangle,
    double=white,
    double distance=0.1em,
    inner sep=0.2em,
    draw=black, very thick,
    minimum height=2.90em,
    minimum width=6.25em,
    text centered,
  },
  point/.style={
    circle,
    inner sep=0pt,
    minimum size=3pt,
    fill=red
  },
  adder/.style={
    circle,
    inner sep=2pt,
    minimum size=0.3in,
    draw=black, very thick,
    text centered
  },
  arrow/.style={
    thick,
    ->,
  >=stealth},
  darrow/.style={
    thick,
    <->,
  >=stealth},
  block/.style={
        draw,
        rectangle,
        rounded corners,
        inner sep=0pt,
        fill=white,
        fill opacity=1.0,
        text opacity=1.0
    }
}

\usepackage{soul}

\usepackage{scalerel}
\usetikzlibrary{svg.path}
\definecolor{orcidlogocol}{HTML}{A6CE39}
\tikzset{
  orcidlogo/.pic={
    \fill[orcidlogocol] svg{M256,128c0,70.7-57.3,128-128,128C57.3,256,0,198.7,0,128C0,57.3,57.3,0,128,0C198.7,0,256,57.3,256,128z};
    \fill[white] svg{M86.3,186.2H70.9V79.1h15.4v48.4V186.2z}
    svg{M108.9,79.1h41.6c39.6,0,57,28.3,57,53.6c0,27.5-21.5,53.6-56.8,53.6h-41.8V79.1z M124.3,172.4h24.5c34.9,0,42.9-26.5,42.9-39.7c0-21.5-13.7-39.7-43.7-39.7h-23.7V172.4z}
    svg{M88.7,56.8c0,5.5-4.5,10.1-10.1,10.1c-5.6,0-10.1-4.6-10.1-10.1c0-5.6,4.5-10.1,10.1-10.1C84.2,46.7,88.7,51.3,88.7,56.8z};
  }
}
\newcommand\orcidicon[1]{\href{https://orcid.org/#1}{\mbox{\scalerel*{
        \begin{tikzpicture}[yscale=-1,transform shape]
          \pic{orcidlogo};
        \end{tikzpicture}
}{|}}}}

\title{
Swarming Without an Anchor (SWA): Robot Swarms Adapt Better to Localization Dropouts Then a Single Robot
}

\author{Ji\v{r}\'{i} Horyna$^{1\orcidicon{0000-0001-6614-0928}}$, 
Roland Jung$^{2\orcidicon{0000-0003-4622-0079}}$, 
Stephan Weiss$^{2\orcidicon{0000-0001-6906-5409}}$,
Eliseo Ferrante$^{3\orcidicon{0000-0002-2213-8356}}$ and
Martin Saska$^{1\orcidicon{0000-0001-7106-3816}}$
  \thanks{Manuscript received: November 28, 2024; Revised February 26, 2025; Accepted April 1, 2025.}
  \thanks{
  This paper was recommended for publication by Editor M. Ani Hsieh upon evaluation of the Associate Editor and Reviewers' comments.
This work was funded by CTU grant no SGS23/177/OHK3/3T/13 and by the Czech Science Foundation (GAČR) under research project no. 23-07517S.
}
  \thanks{$^1$Multi-Robot Systems Group, Faculty of Electrical Engineering, Czech Technical University in Prague, Technicka 2, Prague, Czech Republic, {\tt\footnotesize\{\href{mailto:horynjir@fel.cvut.cz}{horynjir}|\href{mailto:martin.saska@fel.cvut.cz}{martin.saska}\}@fel.cvut.cz}.}
  \thanks{$^{2}$Control of Networked Systems, Institute of Smart Systems Technologies, University of Klagenfurt, Lakeside Park, Klagenfurt, Austria, {\tt\footnotesize\{\href{mailto:roland.jung@aau.at}{roland.jung}|\href{mailto:stephan.weiss@aau.at}{stephan.weiss}\}@aau.at}.}
  \thanks{$^{3a}$Department of Computer Science, Vrije Universiteit Amsterdam, Amsterdam, The Netherlands, $^{3b}$Department of Computer Science, New York University Abu Dhabi, UAE, $^{3c}$Dubai Future Labs, Dubai, UAE, {\tt\footnotesize\href{mailto:e.ferrante@vu.nl}{e.ferrante@vu.nl}}. 
  \newline 
  Digital Object Identifier (DOI): see top of this page.}
}

\begin{document}

\maketitle


\begin{abstract}
  In this paper, we present the Swarming Without an Anchor (SWA) approach to state estimation in swarms of Unmanned Aerial Vehicles (UAVs) experiencing ego-localization dropout, where individual agents are \rev{laterally} stabilized using relative information only.
  We propose to fuse decentralized state estimation with robust mutual perception and onboard sensor data to maintain accurate state awareness despite intermittent localization failures. 
  Thus, the relative information used to estimate the \rev{lateral} state of UAVs enables the identification of the unambiguous state of UAVs with respect to the local constellation.
  The resulting behavior reaches velocity consensus, as this task can be referred to as the double integrator synchronization problem. 
  All disturbances and performance degradations except a uniform translation drift of the swarm as a whole is attenuated which is enabling new opportunities in using tight cooperation for increasing reliability and resilience of multi-UAV systems.
  Simulations and real-world experiments validate the effectiveness of our approach, demonstrating its capability to sustain cohesive swarm behavior in challenging conditions of unreliable or unavailable primary localization. 
\end{abstract}

\begin{IEEEkeywords}
Distributed Robot Systems, Swarm Robotics, Sensor Fusion
\end{IEEEkeywords}


\section{Introduction}
\begin{figure}[t]
  \setlength\belowcaptionskip{-1.3\baselineskip}
  \centering
  \includegraphics[width=0.49\textwidth]{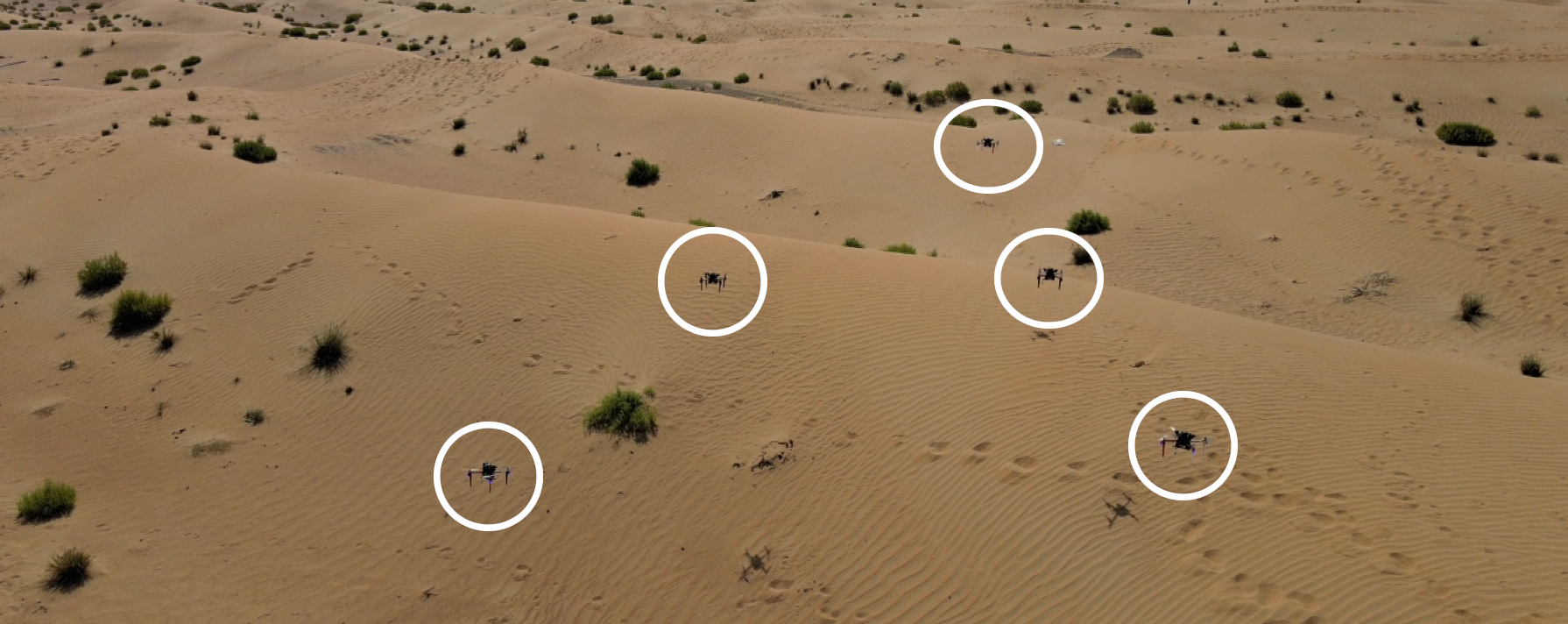}
  \caption{In a feature-poor environment like a flat sandy desert, a multi-UAV system faces challenges as onboard visual or LiDAR localization may not serve as alternatives to GNSS. This scenario requires innovative approaches to maintain accurate estimation despite the lack of features for navigation.}
  \label{fig:intro}
\end{figure}

\rev{
\IEEEPARstart{U}{AV} swarms enhance mission capabilities by leveraging cooperative behavior to perform tasks more efficiently than single UAVs~\cite{horyna2022sar, cardona2019robot, couceiro2013collective, carpentiero2017swarm, albani2018dynamic, radoglou2020compilation, qu2022uav}.
They provide redundancy, enabling continued operation even if individual agents fail. A critical aspect of UAV swarm resilience is reliable state estimation, typically provided by systems such as the Global Navigation Satellite System (GNSS) or Visual Inertial Odometry (VIO). However, when these localization methods fail across all UAVs, standard approaches struggle to maintain stability, often leading to mission failure or crashes.

Swarm resilience to single points of failure can be enhanced through Multi-Robot State Estimation (MRSE) approaches~\cite{horyna2022estimation, horyna2024fast}. In MRSE, agents estimate their states—such as position, velocity, and orientation—by sharing onboard state information, observing neighboring agents through onboard perception systems, and collectively processing sensor data. This method improves system resilience, as the failure or degradation of one robot’s sensors can be compensated for by others, ensuring continuous and reliable state awareness. However, MRSE alone cannot guarantee resilience when all UAVs in the swarm experience simultaneous state estimation degradation, referred to as global degradation of state estimation performance.

To address these limitations, we propose a novel MRSE and control framework that enables UAVs to continue operating even under complete lateral localization subsystem degradation. Our approach utilizes an onboard Mutual Perception System (MPS) to estimate agents’ lateral states (position and its derivatives in the horizontal plane) relative to the local UAV constellation.
Unlike conventional approaches that fail when VIO becomes unreliable—such as over visually uniform terrain (e.g., sand dunes in \autoref{fig:intro}, rivers) or in GNSS-denied environments (e.g., jammed areas, urban canyons, forests)—our framework ensures the swarm remains coordinated despite the loss of absolute positioning.

By enforcing velocity synchronization~\cite{zhang2022active}, the formation experiences gradual drift rather than destabilization, allowing UAVs to traverse degraded regions and resume normal operation once external localization becomes available again.
This purely relative lateral state estimation approach significantly improves swarm resilience compared to single-agent systems and traditional localization-dependent methods.
}

\subsection{Related work}
\rev{
Recent advancements in fail-safe mechanisms for UAVs have significantly enhanced autonomous flight safety and reliability. In \cite{hasan2019model}, a model-based fail-safe module detects potential failures in real-time and deploys a parachute to prevent crashes, proving effective through simulations and real-world tests. The novel visual teach and repeat (VT\&R) system introduced in \cite{warren2018there} enables UAVs to navigate back to their starting point using visual landmarks in GNSS-denied environments. The authors of \cite{sulieman2024path} propose a robust path planning strategy that generates reliable flight paths using onboard sensors to mitigate permanent GNSS failures. Additionally, \cite{ochoa2017fail} focuses on fail-safe mechanisms for urban settings by integrating sensor systems and real-time decision-making algorithms, enhancing navigation and risk mitigation. 
Compared to these works, the proposed approach introduces resilience against global degradation of localization performance among a swarm, allowing the mission to continue.

Research in multi-robot state estimation and cooperative localization, especially in GNSS-denied environments, has progressed significantly but remains limited. Traditional methods involve sharing GNSS, inertial, and Ultra-Wideband (UWB) data among UAVs for localization during GNSS signal loss, as shown in \cite{goel2017distributed, qu2010cooperative1, qu2010cooperative2, qu2011cooperative, wan2014cooperative}. The leader-follower structure explored by \cite{russell2019cooperative} involves a GNSS-equipped UAV aiding another through inertial measurements. Key contributions from \cite{causa2019improving} include protocols enabling UAVs to share sensor data, enhancing autonomous navigation despite GNSS challenges. The work of \cite{horyna2022estimation, horyna2024fast} emphasizes onboard cooperative estimation, validating methods through real-world experiments with UAV swarms. The distributed Kalman filter approach in \cite{roumeliotis2000distributed} improves localization accuracy through relative measurements and odometry data. In contrast, our solution leverages relative information solely for the lateral stabilization of agents, allowing the swarm to operate independently of environmental awareness and outperform single-robot data reliance. \rev{ In \cite{shalaby2024multi}, a multi-robot relative pose estimation method enables drones to estimate their positions by passively listening to UWB signals from neighbors. This approach, which combines distance measurements with IMU data using an Extended Kalman Filter (EKF), may be more vulnerable to communication deficiencies than our method, which includes a tightly coupled control strategy alongside state estimation. }

Formation coordination and state consensus in multi-robot systems have employed various methodologies to address challenges like the lack of velocity measurements. For instance, \cite{zhang2017novel} presents an algorithm for consensus in second-order systems using position measurements and a virtual coupling term to estimate velocities. \cite{zhang2022coordination} develops a robust framework for multi-agent systems facing antagonistic dynamics, ensuring consensus under adversarial interactions. The Decoupled Method in \cite{cui2018sampled} uses sampled-based strategies to maintain convergence in nonlinear multi-agent systems despite deception attacks. \cite{lu2020distributed} proposes a distributed event-triggered control strategy for linear systems to minimize communication while achieving finite-time consensus. \cite{lin2023velocity} introduces a velocity-free formation control method for quadrotors, enabling precise tracking without direct velocity measurements. 
In contrast, our proposed method achieves state consensus among swarm UAVs through the design of lateral state estimation and control systems that rely solely on relative information, enabling a seamless and resilient framework for coordinated state estimation that allows the swarm to maintain cohesion.
}

\subsection{Contribution}
The primary contribution of this work is the introduction of a distributed state estimation and control system SWA capable of \rev{laterally} stabilizing UAVs using only relative position measurements and estimating states expressed in local reference frames. To facilitate this, a novel reference control frame is defined in an unambiguous position within the local constellation. 
This system stabilizes individual UAVs' positions relative to surrounding UAVs' positions, which is particularly advantageous in scenarios involving long-term degradation of global localization performance, where the absence of environmental localization anchors could be devastating and even life-threatening in a non-resilient system.
Additionally, we demonstrate that utilizing relative mutual positions for state estimation and control of individual UAVs results in a uniform drift of the entire formation. This verifies that the proposed state estimation framework functions analogously as a low-pass filter for disturbances affecting the individual UAVs, which is crucial for stabilizing the swarm in zero features situations.

\subsection{Problem statement \& Preliminaries}
This work addresses the \rev{lateral} stabilization of UAVs within a swarm using only relative position measurements, which enhances the overall resilience of the swarm system in the event of global degradation of primary state estimates. \rev{We assume the environment is unknown but obstacle-free (or obstacle detection and avoidance is available), and focus on stabilizing UAVs in the horizontal plane, as lateral kinematics is most vulnerable to common state estimation failures}. Vertical and heading states are estimated using additional sensors in the estimation pipeline.
Horizontal velocity consensus is anticipated to be achieved among swarm members, thereby stabilizing individual UAVs relative to each other within the group. The estimation approach is fully decoupled and independent of estimates from other UAVs. 
\rev{To meet these requirements, the swarm UAVs estimate the relative positions of surrounding agents using an onboard MPS. 
Nevertheless, this representation is interchangeable with separate bearing and range measurements when accounting for noise in the transformation between these representations.}

\rev{We assume that UAVs interact with each other solely through the proposed algorithms. Any other explicit interaction, such as communication, is not permitted.}

\begin{figure}[t]
  \vspace{0.25cm}
  \setlength\belowcaptionskip{-1.3\baselineskip}
  \centering
  \includegraphics[page=1, trim={0.0cm 0.1cm 0.7cm 4.1cm}, width=0.49\textwidth, clip]{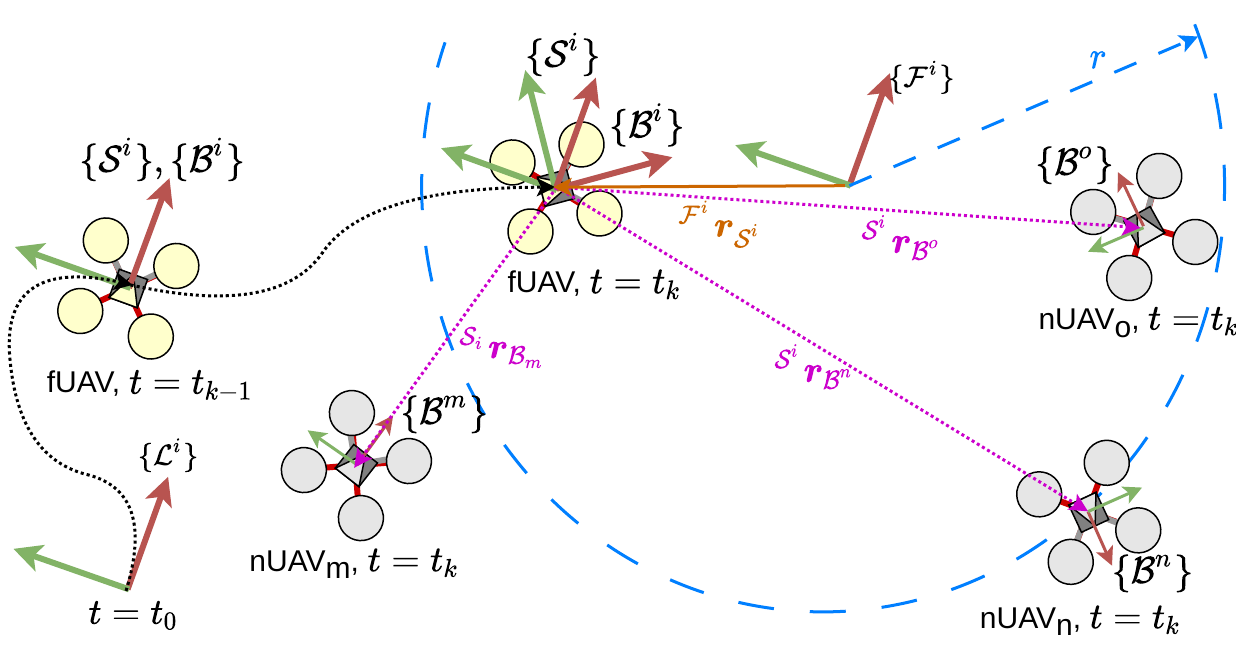}
  \caption{Frames of reference of the fUAV. The x- and y-axis are held in red and green, respectively. }
  \label{fig:frames}
\end{figure}

Most of the mathematical expressions in this work are introduced in the stable frame $\refframe{S^i}$, which is fixed to the center of mass of \emph{i}-th UAV and aligned to the initial pose of the rotational body frame \emph{$\refframe{B^i}$} (see~\autoref{fig:frames}). 
The rotational body frame $\refframe{B^i}$ is fixed to the UAV's center of mass, with x-axis determining the front of the UAV, and z-axis parallel to the negative vector of gravitational field intensity.
Last but not least, the \emph{i}-th UAV is controlled in the floating frame $\refframe{F^i}$, which is parallel to the frame $\refframe{S^i}$ and positioned at the desired position of \emph{i}-th UAV in the swarm. The essence of the frame $\refframe{S^i}$ lies in an effort to estimate neighboring UAVs' states in a stationary frame, and in a straightforward definition of the frame $\refframe{F^i}$, which is crucial for controlling the UAV. 
\rev{The choice of the $F_i$ frame for control and state estimation is primarily practical rather than essential, as its transformation to other frames is straightforward.}
The absolute orientation of the UAV's navigation frame $\refframe{L^i}$ (origin heading at take-off), the stable frame $\refframe{S^i}$ and the floating frame $\refframe{F^i}$ is equal, meaning that $\reference{\vR}{\cL}{\cS}{} = \reference{\vR}{\cS}{\cF}{} = \vI$. Thus, the orientation $\reference{\vR}{\cL}{\cB}{}$ is known and, e.g., provided by a different heading estimator.

\subsection{Terminology \& Notation}
In this work, we employ the following terminology: \emph{Swarm} (S) denotes the set of all cooperating homogeneous UAVs. The \emph{focal UAV} (fUAV) is an arbitrary UAV from S on which the proposed methods are 
\rev{demonstrated. The same algorithms are applied onboard each UAV, establishing a distributed topology.} 
\emph{Surroundings} (O) refers to the subset of S that includes all UAVs observable by the fUAV. An \emph{observable UAV} (oUAV) is any UAV within O. The \emph{neighborhood} (N) is the subset of O that comprises UAVs directly affecting the behavior of the fUAV through the proposed algorithms, identified using a neighborhood selection algorithm from \cite{horyna2022sar}. A \emph{Neighbor} (nUAV) is any UAV within N.

Throughout this paper, the following notation is used. A normally distributed multivariate variable is defined as $\vx^{i} \sim \cN(\hat{\vx}^{i}, \bSigma^{ii})$, with a mean $\hat{\vx}^i$ and covariance (uncertainty) $\bSigma^{ii}$, which is called the \textit{belief} of $i$. The time indices of state variables are indicated by the right \rev{subscript}. E.g., $\vx_{k}$, denotes the state at the time $t({k})$. Names of reference frames are capitalized and calligraphic, e.g., $\refframe{\cB}$ for the rigid body.
A pose between the reference frames $\refframe{\cA}$ and $\refframe{\cB}$ is defined as $\reference{\vT}{\cA}{\cB}{} \in SE(2) \coloneqq \bigg \{\begin{bmatrix} \reference{\vR}{\cA}{\cB}{} & \reference{\vr}{\cA}{\cB}{} \\ \mathbf{0}^\transpose & 1 \end{bmatrix} \bigg | \vR \in SO(2), \rev{\vr} \in \mathbb{R}^2 \bigg \} $ (read as $\reference{\vx}{from}{~to}{}$). \rev{E.g., a point $p$ expressed in frame $\cB$ by $\reference{\vr}{\cB}{p}{}$, can be expressed in $\cA$ by $\begin{bmatrix} \reference{\vr}{\cA}{p}{} \\ 1 \end{bmatrix} =\reference{\vT}{\cA}{\cB}{} \begin{bmatrix} \reference{\vr}{\cB}{p}{} \\ 1 \end{bmatrix}$. The inverse of a rotation matrix is it's transpose $\reference{\vR}{\cA}{\cB}{} = \referencet{\vR}{\cB}{\cA}{}{\transpose}$. $\vI_n$ and $\bZero_n$ are $n \times n$ identity and null matrices, respectively.}

\begin{figure*}[ht]
  \setlength\belowcaptionskip{-1.0\baselineskip}
  \vspace{0.25cm}
  \centering
    \includegraphics[page=1, trim={0.7cm 0.1cm 1.4cm 0.3cm}, width=1.0\textwidth, clip]{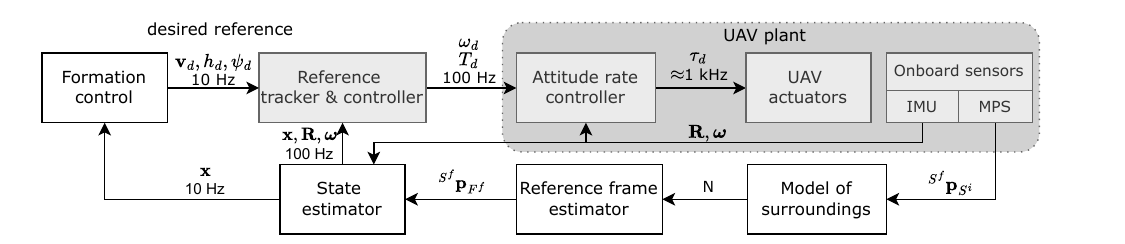}
  \caption{
    A diagram of the system architecture: \emph{Formation control} supplies the desired velocity, height, and heading to a lower control layer \cite{baca2021mrs}. It maintains cohesion with surrounding UAVs and stabilizes the focal UAV in the local constellation. \emph{Model of surroundings} treats oUAVs as point masses via a bank of LKFs. \emph{Reference frame estimator} defines an unambiguous position of the floating frame relative to the oUAVs model. The focal UAV state is estimated in \emph{State estimator} using a single LKF.
  }
  \label{fig:archi}
\end{figure*}

\section{Methodology}
The description of the overall solution of the problem addressed in this paper is divided into several interconnected subsystems, as shown in \autoref{fig:archi}.
The \emph{Model of surroundings} represents the oUAVs as point masses using a bank of Linear Kalman Filters (LKFs), which filter out noisy measurements from the MPS. The \emph{Reference frame estimator} then defines the unambiguous position of the floating reference frame based on the output of the oUAVs model. Ensuring unambiguity at this stage is crucial, as no ground-fixed reference frame exists in the scenario being studied.
The \emph{State estimator} uses a single LKF to estimate the state of the focal UAV, providing this information to the \emph{Formation control}. The \emph{Formation control} module then determines the desired velocity, height, and heading, which are passed to the lower layers of the control pipeline \cite{baca2021mrs}. This module ensures that the UAV remains coordinated with the surrounding UAVs and keeps the focal UAV stable within the group. 

\subsection{Model and state estimation of surroundings}
\label{sec:model}
oUAVs are modeled as point masses capable of motion in the two-dimensional space. 
The \rev{motion} dynamics of \emph{i}-th oUAV is modeled as a discrete and decoupled Linear Time-Invariant (LTI) system assuming a constant acceleration dynamic model without a control input:
\begin{align}
  \label{eq:model}
  \mathbf{x}^{i}_{o,k} =& \mathbf{F}_o\mathbf{x}^i_{k-1} + \mathbf{w}_{o,k}.
\end{align}

The state vector $\mathbf{x}^i_{o,k}$ of the i-th observed UAV at $t=t^k$, the discrete-time state transition matrix $\mathbf{F}_o$, and the process noise $\mathbf{w}_{o,k}$ are defined as:
\begin{align}
  \label{eq:model_info}
  \mathbf{x}_{o,k}^{i} =& \left[
    \begin{smallmatrix}
      \reference{\vr}{\cS^f}{\cS^i}{\cS^f} \\
      \reference{\dot{\vr}}{\cS^f}{\cS^i}{\cS^f} \\
      \reference{\ddot{\vr}}{\cS^f}{\cS^i}{\cS^f} \\
    \end{smallmatrix}
    \right],
  \rev{\mathbf{F}_o = \left[
    \begin{smallmatrix}
      \vI_2 & \vI_2 \Delta t & \vI_2 \frac{\Delta t^2}{2} \\
      \bZero_2 & \vI_2 & \vI_2\Delta t \\
      \bZero_2 & \bZero_2 & \vI_2 \\
    \end{smallmatrix}
    \right]},\\
  \mathbf{w}_{o,k} \sim& (\mathbf{0}, \mathbf{Q}_o), 
\end{align}
where $\Delta t$ is time between two consequent computational steps $k$ and $k-1$, and $\mathbf{Q}_o$ is a diagonal process noise covariance matrix with dimensions of $\mathbf{F}_o$. The state vector comprises the positon, velocity and acceleration of the \emph{i}-th oUAVs with respect to the stable frame of the focal UAV. The measurement model of the system is defined as follows:
\begin{align}
  \label{eq:meas_model}
  \vz^i_{k} =& \mathbf{H}_o \vx^i_{o,k} + \vv_{o,k},
\end{align}
where $\mathbf{H}_o$ is measurement matrix and $\mathbf{v}_{o,k} \sim (\mathbf{0}, \mathbf{R}_o)$ is measurement noise described by covariance matrix $\mathbf{R}_o$. Measurements of the states are obtained as the relative positions of oUAVs estimated by onboard MPS with respect to the body frame $\reference{\vr}{\cB^{f}}{\cS^{i}}{}, i \in \text{O}$, which is transformed into \emph{$S^{f}$} frame with a known rotation between the \emph{$B^f$}  and \emph{$S^f$} frames:
\begin{align}
  \label{eq:measurements2}
  \rev{
  \mathbf{\vz}_{k}^{i} = \reference{\vr}{\cS_{k}^{f}}{\cS_{k}^{i}}{} = \referencet{\vR}{\cB_{k}^{f}}{\cS_{k}^{f}}{}{\transpose} \reference{\vr}{\cB_{k}^{f}}{\cS_{k}^{i}}{}.
  }
\end{align}

\rev{Thus, the measurement matrix is in the form: 
\begin{align}
  \label{eq:meas_matrix}
  \mathbf{H}_o = \left[
      \vI_2 ~ \bZero_{2 \times 4}
    \right].
\end{align}}
The proposed filtration and state estimation approach uses one LKF for each neighbor UAV, forming a bank of LKFs. This method provides greater flexibility, scalability and adaptability compared to using a single filter to estimate the states of all oUAVs.
A new LKF is added to the bank when a new UAV is detected, while an LKF is removed from the bank if the UAV is no longer detected for a specified time threshold.
\rev{The bank of LKFs is managed in C++ as a vector of custom structures, each storing all available data about oUAVs, including their ID, an instance of the LKF filter, timestamps of the last corrections and updates, and other relevant information.}
The \rev{state estimate} $\hat{\mathbf{x}}^i = \hat{\mathbf{x}}^{i}_{o}$ and covariance $\mathbf{P}^i$ of \emph{i}-th filter's state are updated according to:
\begin{align}
  \label{eq:corr1}
  \mathbf{K}^i_k &= \mathbf{P}^i_{k|k-1} \mathbf{H}^T_o (\mathbf{H}_o \mathbf{P}^i_{k|k-1} \mathbf{H}^T_o + \mathbf{R}_o)^{-1}, \\
  \label{eq:corr2}
  \hat{\mathbf{x}}^i_{k|k} &= \hat{\mathbf{x}}^i_{k|k-1} + \mathbf{K}^i_k (\mathbf{z}^i_k - \mathbf{H}_o \hat{\mathbf{x}}^i_{k|k-1}), \\
  \label{eq:corr3}
  \mathbf{P}^i_{k|k} &= \mathbf{P}^i_{k|k-1} - \mathbf{K}_k^i \mathbf{H}_o \mathbf{P}^i_{k|k-1},
\end{align}
\rev{with the Kalman gain $\vK$, and the \emph{a priori} and \emph{a posteriori} of the mean and covariance, indicated by $\{\bullet\}_{k|k-1}$ and $\{\bullet\}_{k|k}$, respectively.}
The prediction step of LKFs describes the evolution of state estimate and state covariance in time using the model of the system with previous state estimate and process noise:
\begin{align}
  \label{eq:pred}
  \hat{\mathbf{x}}^i_{k|k-1} &= \mathbf{F}_o \hat{\mathbf{x}}^i_{k-1|k-1}, \\
  \hat{\mathbf{P}}^i_{k|k-1} &= \mathbf{F}_o\mathbf{P}^i_{k-1|k-1}\mathbf{F}^T_o + \mathbf{Q}_o.
\end{align}
Values of matrices $\mathbf{Q}_o$ and $\mathbf{R}_o$ were set to fit the data obtained in statistically processed realistic simulations. 

\subsection{Definition of the floating reference frame}
\label{sec:floating_reference}
In our study case, the floating reference frame $\refframe{F}$ is not fixed to the environment. We assume the fUAV can be stabilized in frame $\refframe{F}$ using only relative observations of neighbors. Without loss of generality, we define $\refframe{F}$ at the desired position of the fUAV in the formation with orientation identical to $\refframe{S}$. \rev{In general, when the number of nUAVs $n > 2$}, we define the desired fUAV position as the center of a circle given by nUAV locations.
The relation between the position of the floating frame and the location of nUAVs is given by:
\begin{align}
  \label{eq:recircle}
  ax_j + by_j + c = x^2_j + y^2_j - r^2,
\end{align}
  where $a = 2x_c$, $b = 2y_c$, $c = -{x_c}^2 -{y_c}^2$, \rev{$r$ is a fixed parameter representing the radius of the circle,} $\mathbf{r_j} = \left[x_j, y_j \right]^T$ are x and y estimated relative coordinates of \emph{j}-th nUAV \rev{$\reference{\vr}{\cS^f}{\cS^j}{}$}, and \rev{$\mathbf{r}_c = \reference{\vr}{\cS^{f}}{\cF}{} = \left[x_c, y_c\right]^T$} are unknown coordinates of the center of the circle. The whole system for $j \in \{ 1, 2, ..., n \}$, where $n$ is number of nUAVs, can be rewritten in the matrix form as:
\begin{align}
  \label{eq:model2}
  \mathbf{A}_c \mathbf{x}_c =& \mathbf{B}_c,
\end{align}
where:
\begin{align}
  \label{eq:lineq}
  \mathbf{A}_c = \left[
    \begin{smallmatrix}
      x_1 & y_1 & 1 \\
      x_2 & y_2 & 1 \\
      \vdots & \vdots & \vdots \\
      x_n & y_n & 1 \\
    \end{smallmatrix}
    \right],
  \mathbf{x}_c = \left[
    \begin{smallmatrix}
      a \\
      b \\
      c \\
    \end{smallmatrix}
    \right],
  \mathbf{B}_c = \left[
    \begin{smallmatrix}
      x^2_1 + y^2_1 - r^2 \\
      x^2_2 + y^2_2 - r^2 \\
      \vdots \\
      x^2_n + y^2_n - r^2 \\
    \end{smallmatrix}
    \right],
\end{align}
The solution of the system is given by:
\begin{align}
  \label{eq:circle-center-solution}
  \hat{\mathbf{x}}_c = \mathbf{A}^{+}_c\mathbf{B}_c = \left( \mathbf{A}^T_c\mathbf{A}_c \right)^{-1}\mathbf{A}^T_c\mathbf{B}_c,
\end{align}
where $\mathbf{A}^{+}_c$ is the pseudoinverse of $\mathbf{A}_c$. The unknown coordinates of the center of the circle are then obtained as $x_c = \frac{a}{2}$ and $y_c = \frac{b}{2}$. 
\rev{In the special case where $n = 2$, there exist two possible solutions, $\mathbf{r}_{c,1}$ and $\mathbf{r}_{c,2}$. In this scenario, the solution closer to the fUAV is selected as the position of the floating reference frame. When $n = 1$, an additional constraint is imposed, requiring the solution to lie on the line connecting the fUAV and the nUAV.}

\vspace{-0.2cm}
\subsection{Estimation of focal UAV's lateral \rev{kinematics} states}
\label{sec:focest}
The fUAV is modeled as a point mass capable of motion in a two-dimensional space, similar to the method introduced in \ref{sec:model}. We assume that the fUAV moves with constant acceleration and that a velocity control input is provided. This simplified motion model does not impact the overall performance of the state estimator due to the high-rate periodic corrections applied by the LKF within the state estimator. The \rev{dynamics} of the fUAV system are modeled as follows:

\begin{align}
  \label{eq:focal_model}
  \mathbf{x}_{f,k} =& \mathbf{F}_f\mathbf{x}_{f,k-1} + \mathbf{B}_f\mathbf{u}_{f,k} + \mathbf{w}_{f, k}.
\end{align}

The state vector $\mathbf{x}_{f,k}$ of the focal UAV at $t=t^k$, the input vector $\mathbf{u}_{f,k}$ state transition matrix $\mathbf{F}_f$, the control input matrix $\mathbf{B}_f$, and the process noise $\mathbf{w}_{f, k}$ are defined as: 
\begin{align}
  \label{eq:focal_model2}
  \mathbf{x}_{f,k} =& \left[
    \begin{smallmatrix}
      \reference{\vr}{\cF}{\cS^f}{} \\
      \reference{\dot{\vr}}{\cF}{\cS^f}{} \\
      \reference{\ddot{\vr}}{\cF}{\cS^f}{} \\
    \end{smallmatrix}
    \right],
  \mathbf{u}_{f,k} = \mathbf{v}_{d, k}, 
  \rev{\mathbf{F}_f = \left[
    \begin{smallmatrix}
      \vI_2 & \vI_2 \Delta t & \vI_2 \frac{\Delta t^2}{2} \\
      \bZero_2 & \vI_2 e_d & \vI_2 \Delta t \\
      \bZero_2  & \bZero_2  & \vI_2 \\
    \end{smallmatrix}
    \right],} \\
  \mathbf{B}_f =& \left[
    \begin{smallmatrix}
      \bZero_2 \\
      \vI_2 (1 - e_d) \\
      \bZero_2 \\
    \end{smallmatrix}
    \right],
  \mathbf{w}_{f,k} \sim (\mathbf{0}, \mathbf{Q}_f),
    e_d = e^{-\frac{\Delta t}{\tau}}, 
\end{align}
where $e_d$ is the damping factor of the environment, $\tau$ is the time constant, and $\mathbf{v}_{d, k}$ is the desired velocity of the fUAV at time step $k$.
Position measurements $\vz^p_{f,k}$ are obtained by solving eq. \eqref{eq:circle-center-solution}.
The measurement model $\vz_{f,k} = \mathbf{H}_f\mathbf{x}_{f,k} + \mathbf{v}_{f,k}$ is used with unsynchronized measurements 
\begin{align}
  \label{eq:measurements}
  \vz^p_{f,k} =& \left[
    \begin{smallmatrix}
      -x_c \\
      -y_c \\
    \end{smallmatrix}
    \right] + \vv_c,
  \vz^a_{f,k} = \left[
    \begin{smallmatrix}
      \ddot{x}_{I} \\
      \ddot{y}_{I} \\
    \end{smallmatrix}
    \right]  + \vv_a,
\end{align}
where $\ddot{x}_{I}, \ddot{y}_I$ are accelerations measured by the IMU unit $\refframe{I}$ projected into the frame $\refframe{S^f}$ and with $\mathbf{v}_{c} \sim (\mathbf{0}, \mathbf{R}_c)$ and $\mathbf{v}_{a} \sim (\mathbf{0}, \mathbf{R}_a)$.

\rev{
The proposed approach utilizes pre-filtered IMU data by transforming the estimated UAV’s tilt into an estimated lateral accelerations according to the equation:
\begin{align}
   \left[
    \begin{smallmatrix}
      \ddot{x}_{I} \\
      \ddot{y}_{I} \\
    \end{smallmatrix}
    \right]
    =
    q_c \left[
    \begin{smallmatrix}
      \cos{\psi} & - \sin{\psi} \\
      \sin{\psi} & \cos{\psi}\\
    \end{smallmatrix}
    \right]
    \left[
    \begin{smallmatrix}
      \theta\\
      \phi\\
    \end{smallmatrix}
    \right],
\end{align}
where $\theta, \phi, \psi$ are pitch, roll, and heading angles respectively, and $q_c$ is a conversion factor describing the relationship between tilt angle and estimated UAV acceleration. The system’s collective nature helps mitigate the influence of IMU biases through relative position measurements and the appropriate choice of measurement and process noise matrices.
}

The measurement matrix $\mathbf{H}^f$ is defined as: 
\begin{align}
  \label{eq:meas_matrix2}
  \mathbf{H}_f = \rev{\left[
       h_p \vI_2 ~ \bZero_2 ~  h_a \vI_2
    \right]}, h_p, h_a \in \{ 0, 1 \}, h_p \neq h_a,
\end{align}
where $h_p = 1$, when the position measurement is obtained, and $h_a = 1$, when the acceleration measurement is obtained.

The prediction step of the fUAV's state estimate is:
\begin{align}
  \label{eq:pred2}
  \hat{\mathbf{x}}_{f,k|k-1} &= \mathbf{F}_f \hat{\mathbf{x}}_{f,k-1|k-1} + \mathbf{B}_f\mathbf{u}_{f,k}, \\
  \hat{\mathbf{P}}_{f,k|k-1} &= \mathbf{F}_f\mathbf{P}_{f,k-1|k-1}\mathbf{F}^T_f + \mathbf{Q}_f.
\end{align}
A correction step equivalent to eq.~\eqref{eq:corr1} - \eqref{eq:corr3} is applied to update the state estimate and covariance.
Values of matrices $\mathbf{Q}_f$, $\mathbf{R}_c$ and $\mathbf{R}_v$ were set to fit the data obtained in statistically processed realistic simulations. 

\subsection{Distributed formation control}
We propose a control law based on a state feedback controller defining desired lateral velocity of the fUAV $\mathbf{v}_d$ as 
\begin{equation}
  \label{eq:control}
  \begin{split}
    \mathbf{v}_d = &\overbrace{-k_p \mathbf{e}_p}^{\begin{tabular}{c}
      \tiny position feedback
    \end{tabular}} + \overbrace{-k_v \mathbf{e}_v}^{\begin{tabular}{c}
      \tiny velocity feedback
    \end{tabular}},
  \end{split}
\end{equation}
where $\mathbf{e}_p =\rev{ - \hat{\mathbf{x}}_{f,k}\left[\mathbf{I}_2 ~ \mathbf{0}_2 ~ \mathbf{0}_2 \right]}$, $\mathbf{e}_v =\rev{ - \hat{\mathbf{x}}_{f,k}\left[\mathbf{0}_2 ~ \mathbf{I}_2 ~ \mathbf{0}_2 \right]}$ are position and velocity control errors, and $k_p$, $k_v$ are position and velocity feedback gains. The control errors correspond to the negative relative state of the fUAV in the frame \emph{$F$}. 

\subsection{Observability of the swarm system}
\begin{figure}[t]
  \vspace{0.25cm}
  \setlength\belowcaptionskip{-1.3\baselineskip}
  \centering
  \includegraphics[page=1, trim={0.2cm 0.2cm 3.5cm 0.0cm}, width=0.49\textwidth, clip]{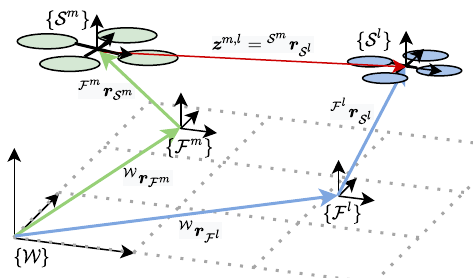}
  \caption{Frames of reference of the combined swarm system. The red line indicates a relative position measurement between the UAV $m$ and $l$.}
  \label{fig:frames2}
\end{figure}

Determining if a system relying solely on relative position measurements is jointly observable involves assessing the observability of the swarm state space across all interconnected subsystems of the agents. We consider only a horizontal subset of the UAVs' states, which are mapped onto the world reference. As shown in \autoref{fig:frames2}, the combined system can be represented by state-space equations, which include a state vector and a state transition matrix:
\begin{align}
  \label{eq:obs1}
  \reference{\vr}{\cW}{\cS^i}{}{i} = \remove{\mathbf{r}^i =}
  \left[
    \begin{smallmatrix}
      x^i \\
      y^i \\
    \end{smallmatrix}
    \right] 
  = \reference{\vr}{\cW}{\cF^i}{} + \reference{\vR}{\cW}{\cF^i}{}  \reference{\vr}{\cF^i}{\cS^{f,i}}{}, \\
  \mathbf{x} = \left[
    \begin{smallmatrix}
      \cdots & x^i & y^i & \dot{x}^i & \dot{y}^i & \cdots 
    \end{smallmatrix}
    \right]^T,\mathbf{F}_c = \left[
    \begin{smallmatrix}
      \ddots & \mathbf{0} & \mathbf{0}\\
      \mathbf{0} & \begin{smallmatrix}
        \vI_2 & \vI_2 \Delta t \\
        \bZero_2 & \vI_2  \\
      \end{smallmatrix} & \mathbf{0} \\
      \mathbf{0} & \mathbf{0} & \ddots\\
    \end{smallmatrix}
    \right],
\end{align}
with $\reference{\vR}{\cW}{\cF^i}{} = \vI$ (is assumed to be known). 
If we consider relative mutual position measurements only, the measurement subsystem can be defined as:
\begin{align}
  \label{eq:obs2}
  \mathbf{z} &= \mathbf{H}_c\mathbf{x};\, 
  \mathbf{z} = \left[
    \begin{smallmatrix}
      \vdots \\
      \mathbf{z}^{kl} \\
      \vdots \\
    \end{smallmatrix}
    \right],\\
  \mathbf{z}^{kl} &= \left[
    \begin{smallmatrix}
      z^{kl}_x \\
      z^{kl}_y \\
    \end{smallmatrix}
    \right]
 =  \reference{\vr}{\cW}{\cF^k}{} + \reference{\vr}{\cF^k}{\cS^k}{} - \reference{\vr}{\cW}{\cF^l}{} - \reference{\vr}{\cF^l}{\cS^l}{}.
\end{align}
\rev{The measurement matrix $\mathbf{H}_c$ for $n$ UAVs is a matrix with $n(n-1)$ rows and $4n$ columns, where each row corresponds to a relative position measurement between a pair of UAVs, and each entry reflects the difference between the relevant state variables for those UAVs. Please note that each UAV has its own focal frame.}

The rank of the observability matrix of the combined system is  
\begin{align}
  \label{eq:obs3}
rank(\mathcal{O}) = dim(\mathbf{x}) - 4 = \rev{4n - 4},\end{align} 
where
\begin{align}
  \label{eq:obs4}
  \mathcal{O} = \left[
    \begin{smallmatrix}
      \mathbf{F}_c \\
      \mathbf{F}_c\mathbf{H}_c \\
      \vdots \\
      \mathbf{F}_c\mathbf{H}_c^{\rev{4n-4}} \\
    \end{smallmatrix}
    \right],
\end{align} 
with the basis of unobservable subspaces:
\begin{align}
  \label{eq:obs5}
  \{\mathbf{1}_n^T \otimes
  \left[
    \begin{smallmatrix}
      1 \\
      0 \\
      0 \\
      0 \\
    \end{smallmatrix}
    \right],\mathbf{1}_n^T \otimes
  \left[
    \begin{smallmatrix}
      0 \\
      1 \\
      0 \\
      0 \\
    \end{smallmatrix}
    \right],\mathbf{1}_n^T \otimes
  \left[
    \begin{smallmatrix}
      0 \\
      0 \\
      1 \\
      0 \\
    \end{smallmatrix}
    \right],\mathbf{1}_n^T \otimes
  \left[
    \begin{smallmatrix}
      0 \\
      0 \\
      0 \\
      1 \\
    \end{smallmatrix}
    \right]\}.
\end{align}
\rev{The expression represents four column vectors, each of size $4n \times 1$, obtained through the Kronecker product operation. The notation $\mathbf{1}_n$ denotes an $n \times 1$ column vector of ones.}

\begin{figure}[t]
  \vspace{0.25cm}
  \setlength\belowcaptionskip{-1.3\baselineskip}
  \centering
  \includegraphics[page=1, trim={1.0cm 0.1cm 2.0cm 1.5cm}, width=0.49\textwidth, clip]{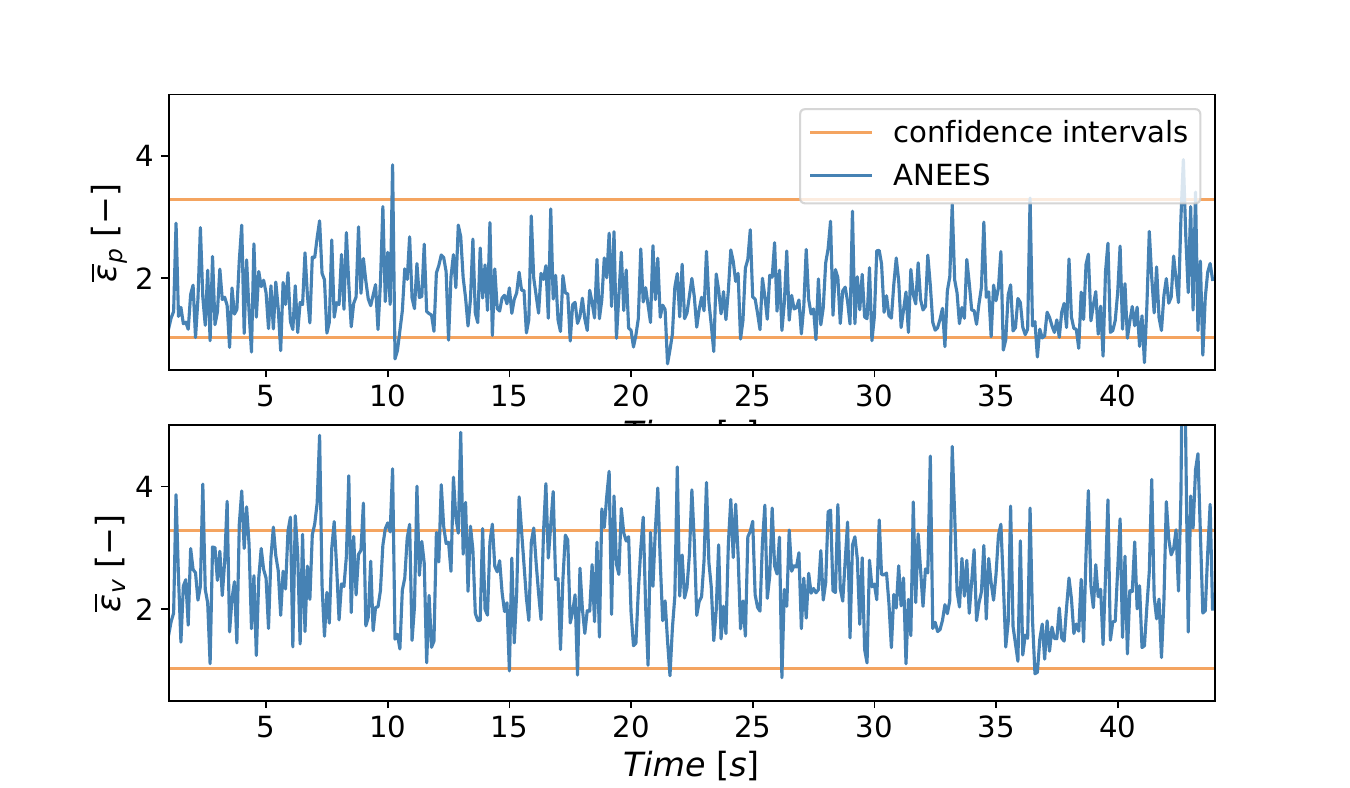}
  \caption{ANEES analyzes indicates consistency of the proposed state estimator.}
  \label{fig:nees}
\end{figure}

\begin{figure*}[t]
  \vspace{0.25cm}
  \begin{subfigure}[t]{1.0\textwidth}
    \centering
    \includegraphics[width=0.99\textwidth]{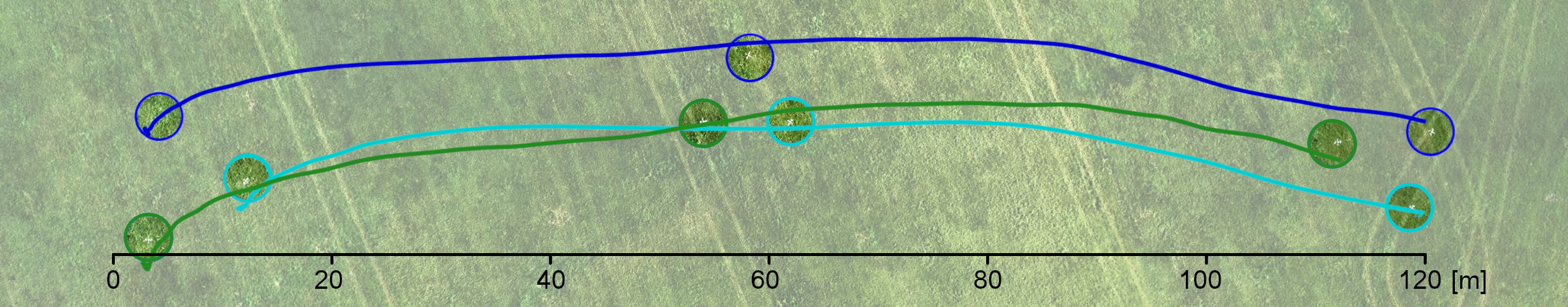}
    \caption{Positions of UAVs during the drift}
    \label{fig:rw2}
  \end{subfigure}
  
  \hfill
  \begin{subfigure}[t]{.445\textwidth}
    \centering
    \includegraphics[page=1, trim={3.2cm 0.7cm 0.9cm 2.15cm}, width=1.0\textwidth, clip]{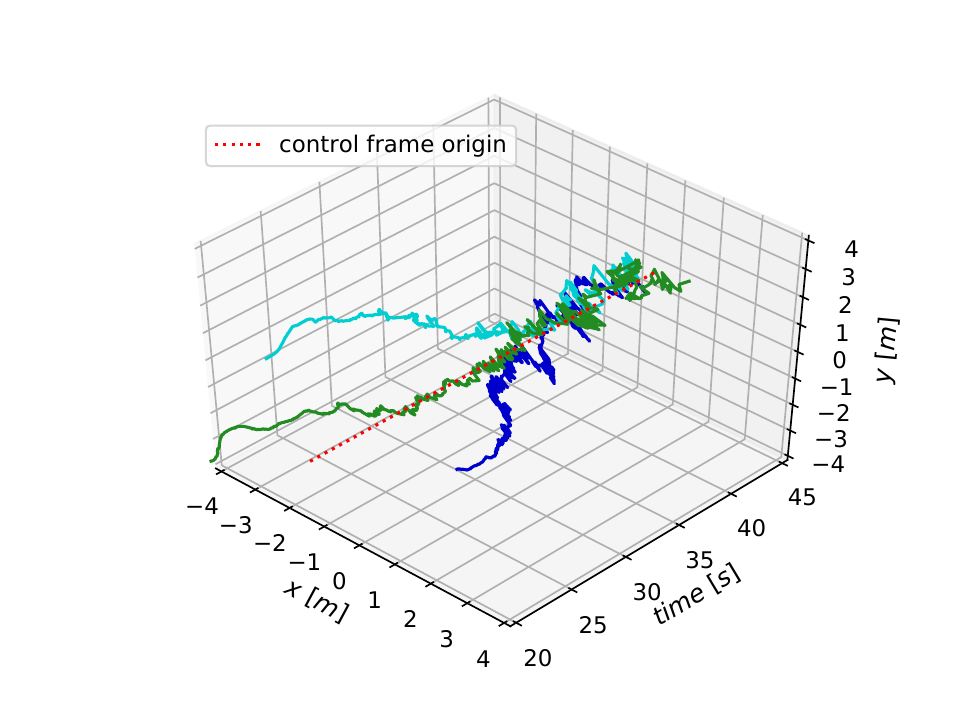}
    \caption{Positions of UAVs in their control frames. UAVs converge to the desired positions relative to the local constellation.}
    \label{fig:rw1}
  \end{subfigure}
  \begin{subfigure}[t]{.545\textwidth}
    \centering
    \includegraphics[page=1, trim={1.55cm 0.1cm 2.0cm 1.5cm}, width=1.0\textwidth, clip]{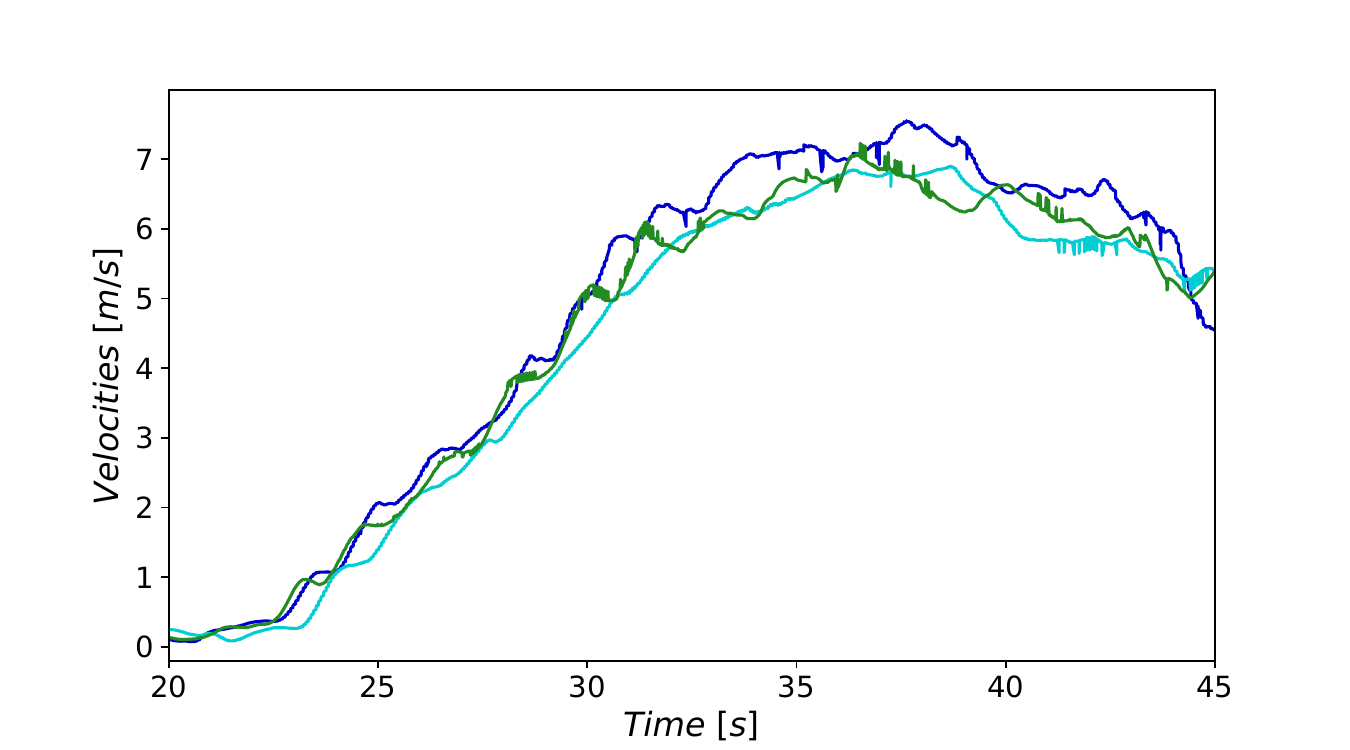}
    \caption{Ground velocities of UAVs.}
    \label{fig:rw3}
  \end{subfigure}
  \label{fig:rw}
  \caption{Experiment with 3 UAVs in the event of full localization dropout. Swarm drifts, but the constellation is remained. Video is available at: \url{https://mrs.fel.cvut.cz/ral-2025-noanchor}.}
\end{figure*}

The unobservable modes correspond to the overall lateral translation of all robots together in the same direction. 
This occurs because measuring only relative positions does not provide any information about the absolute position and velocity of the entire system, leading to four unobservable directions. 
As a result, any uniform shift of the entire system in space cannot be detected through the available measurements.
If additional data, such as the absolute positions of any robot is available, the observability matrix could potentially achieve full rank, making the system fully observable. 
This indicates that the observability problem is more a consequence of the nature of the measurements than an inherent limitation of the system dynamics.

\rev{Although the observability constraints arise from the nature of the measurements}, this insight is significant, as it suggests that increasing the number of swarm agents enhances the system's robustness against a single point of failure. The likelihood of obtaining additional measurements (e.g., camera detection of a distinctive object like a tuft of grass in a visually uniform desert environment) increases with the number of drones in the swarm.

\subsection{Consistency of the state estimate}
When tuning the state estimator for the focal UAV (see \autoref{sec:focest}), we employed the Average Normalized Estimation Error Squared (ANEES) metric to adjust \rev{values of $\mathbf{Q}_f$, $\mathbf{R}_c$ and $\mathbf{R}_v$}, ensuring that the estimator remains unbiased and that it accurately reflects the system's uncertainties. This fine-tuning process enhances the estimator's accuracy and consistency, leading to reliable performance across various scenarios. 
\rev{The simulated scenario consisted of a group of UAVs hovering while a single UAV utilized the proposed state estimation. This UAV was commanded to follow a set of trajectories relative to the rest of the group.}
The ANEES, denoted as $\overline{\epsilon}_k$, is defined as
\begin{align}
  \label{eq:anees1}
  \overline{\epsilon}_k = \frac{1}{K}\sum_{i=1}^K\epsilon_{i,k},
\end{align}
where $K$ is the number of independent data samples, and $\epsilon_k$ is the Normalized Estimation Error Squared (NEES) given by:
\begin{align}
  \label{eq:anees2}
  \epsilon_k = \mathbf{e}_{k|k}^T\mathbf{P}_{k|k}^{-1}\mathbf{e}_{k|k},
\end{align}
where $\mathbf{e}$ is the estimation error and $\mathbf{P}$ is the error covariance matrix. To evaluate the consistency of the state estimate, the acceptance interval $\left[ r_1, r_2 \right]$ is determined as
\begin{align}
  \label{eq:anees3}
  r_1 &= \frac{1}{K}\mathbf{C}^{-1}(\frac{\alpha}{2}, Kn_x),\\
  r_2 &= \frac{1}{K}\mathbf{C}^{-1}(1- \frac{\alpha}{2}, Kn_x),
\end{align}
where $\mathbf{C}^{-1}$ is the inverse cumulative distribution function of the chi-square distribution, $\alpha$ is a parameter defining the confidence region, and $n_x = dim(\mathbf{e})$. The state estimator is considered consistent if
\begin{align}
  \label{eq:anees4}
  \overline{\epsilon}_k \in \left[ r_1, r_2 \right].
\end{align}
To gain deeper insight into the performance of the state estimator, we separately analyzed the position and velocity states. \autoref{fig:nees} illustrates the ANEES for position estimates $\overline{\epsilon}_p$, and velocity estimates $\overline{\epsilon}_v$, using $K = 12$, $n_x = 2$, and $\alpha = 0.05$, which define the boundaries of a commonly used 95\% confidence region. 
The evaluation of the ANEES analysis (\autoref{fig:nees}) indicates strong performance. The ANEES values for both position and velocity estimates consistently fell within the acceptance interval, demonstrating that the estimator performs well in terms of accuracy and reliability.

\section{RESULTS}
\label{sec:res}

\begin{table*}[t]
  \footnotesize
  \caption{\rev{Algorithm parameters from the experiments. $f_p$ and $f_a$ are the frequencies of measurements $z^p$ and $z^a$.}}
  \centering
  \setlength{\tabcolsep}{0.6em} 
{\renewcommand{\arraystretch}{1.25}
  \begin{tabular}{c|cccccccccccc}
    \textbf{Parameter} & $r$ & $\tau$ & $k_p$ & $k_v$ & $Q_o$ & $R_o$ & $Q_f$ & $R_c$ & $R_a$ & $q_c$ & $f_a$ & $f_p$ \\ \hline
    \textbf{Value} & $10.0$ & $2.8$ & $0.5$ & $0.63$ & $diag(\frac{1}{10^{3}}\mathbf{I}_2, \frac{1}{10^{2}}\mathbf{I}_2, \frac{1}{10}\mathbf{I}_2)$ & $\frac{1}{10}\mathbf{I}_2$ & $diag(\frac{5}{10^{2}}\mathbf{I}_2, \frac{5}{10}\mathbf{I}_2, 5\mathbf{I}_2)$ & $\frac{2}{10^{2}}\mathbf{I}_2$  & $2.25\mathbf{I}_2$ & $6.35$ & $100$ & $10$ \\
    \textbf{Unit} & $m$ & $s$ & $s^{-1}$ & $-$ & $m^2, m^2s^{-2}, m^2s^{-4}$ & $m^2$ & $m^2, m^2s^{-2}, m^2s^{-4}$ & $m^2$  & $m^2s^{-4}$ & $m.s^{-1}$ & $Hz$ & $Hz$ \\
  \end{tabular}}
  \label{tab:param}
\end{table*}


Following the preliminary tests in realistic Gazebo simulations within the Robot Operating System (ROS) environment, the proposed estimation approach was validated through a series of real-world experiments. These experiments were conducted using customized quadrotor platforms based on the DJI F450 frame, equipped with a Pixhawk 4 flight control unit and an Intel NUC10i7FNK onboard computer, as detailed in \cite{HertJINTHW_paper}. 
The UAVs were specifically outfitted with a rangefinder for height above ground estimation and a GNSS receiver. The heading of each individual agent was estimated using magnetometer data. While GNSS data were primarily used for localization during system initialization and takeoff, during flight, the shared GNSS data emulated relative position estimation but was not integrated into the estimation and control loop. 
\rev{To preserve realistic conditions of the real world, the relative positions obtained from shared GNSS data were modulated to correspond to the uncertainty and noise characteristics of 3D LiDARs \cite{vrba2023onboard}} but any other MPS capable of estimating the bearing and range of surrounding UAVs such as UVDAR \cite{walter2019uvdar} could be used. 
\rev{
The decision to utilize emulated relative position measurements from shared GNSS data, augmented with noise, was driven by the need to analyze swarm behavior under complete loss of primary localization while minimizing the influence of sensor limitations. This approach allowed for a simplified system design, facilitating a focused evaluation of the algorithm's performance. Additionally, it mitigated risks associated with potential damage to costly sensors, such as 3D LiDARs, during the experimental procedures.
}
The underlying control pipeline deployed onboard the UAVs is elaborated in \cite{baca2021mrs}. \rev{The parameters used during the experiments are summarized in \autoref{tab:param}.}

\subsection{Real-world experiments}
A series of real-world experiments were conducted with three UAVs (results from one experimental run are shown in \autoref{fig:rw1}) that were hovering at a desired height above the ground, using GNSS data for self-localization. 
At a specific point, the onboard state estimators of all UAVs were simultaneously switched to the proposed state estimation approach.
This change simulated a threat of GNSS jamming, rendering the UAVs unable to access position and velocity measurements in the world frame.

The swarm drifted \SI{100}{\meter} (\autoref{fig:rw2}), with a maximum drift velocity of \SI{7}{\meter\per\second} (\autoref{fig:rw3}). This maximum primarily depends on IMU noise distribution, swarm size, and localization precision. Despite challenging conditions, the swarm completed the experiment without collisions, maintaining cohesion among the UAVs using only relative position measurements. \autoref{fig:rw1} shows individual UAV positions over time relative to the floating frame, demonstrating convergence to the frame's origin, which aligns with the intended design of the state estimator and feedback control approach.

\begin{figure}[h]
  \vspace{0.25cm}
  \centering
  \setlength\belowcaptionskip{-1.3\baselineskip}
  \includegraphics[page=1, trim={1.1cm 0.2cm 2.25cm 1.6cm}, width=0.49\textwidth, clip]{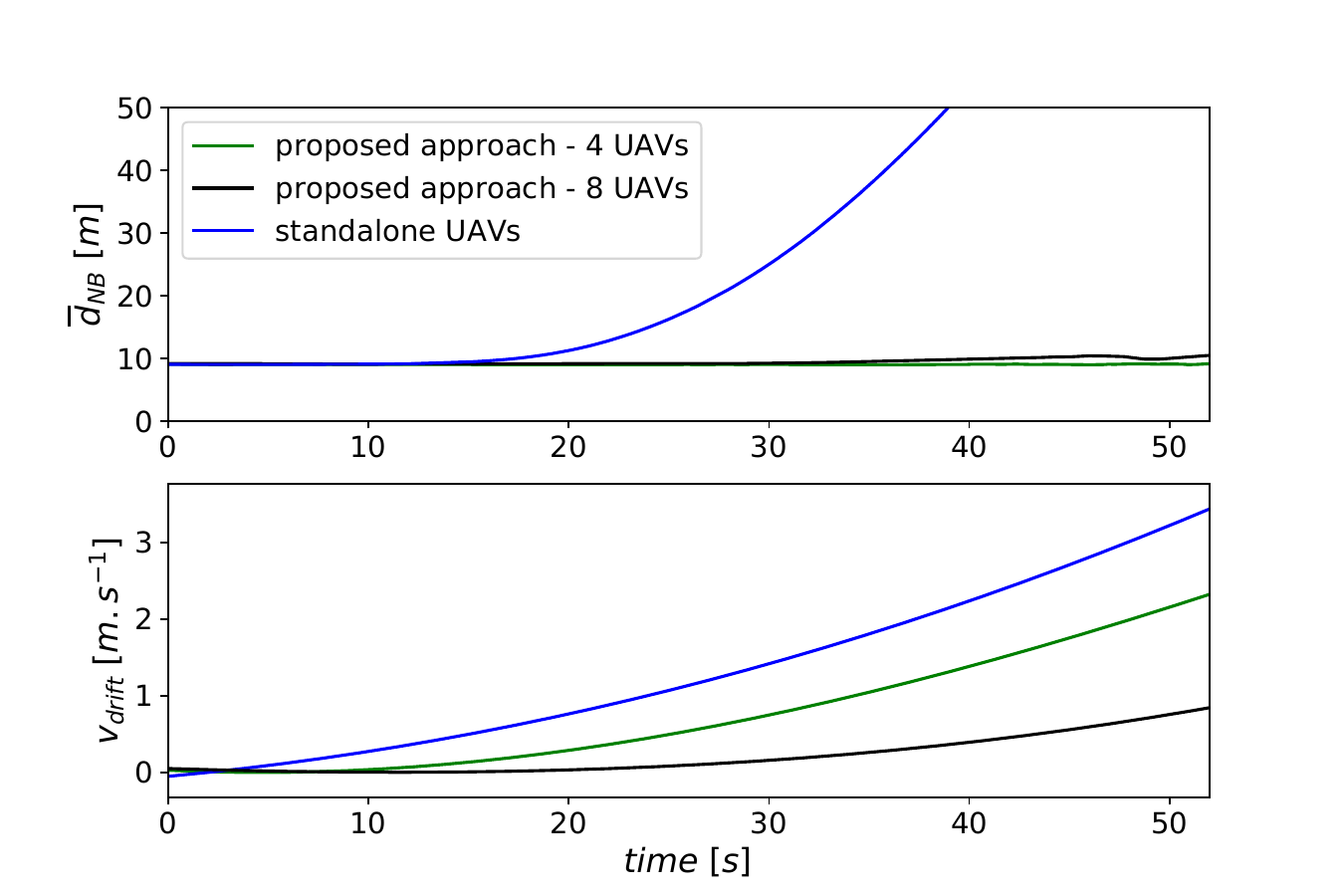}
  \caption{Comparison with standalone UAVs: The swarm employing the proposed approach is capable of maintaining a formation even after localization is lost. Additionally, our results indicate that an increase in the number of UAVs positively influences the stability of the formation, reducing drift.}
  \label{fig:sim}
\end{figure}
\subsection{Simulation - Comparison with a standalone UAVs}

We present a comparison between the proposed estimation approach and a standard multi-robot UAV system without proactive distributed state estimation enhancement (\autoref{fig:sim}). 
\rev{These UAVs possess the same computational capabilities and sensor equipment as the UAV platforms described in \autoref{sec:res}; however, the proposed distributed state estimation system was not implemented on them.
As a baseline, they use the state estimation and control pipeline from~\cite{baca2021mrs}.}
The comparison was conducted through experiments in a realistic Gazebo simulator to mitigate the risk of failures and crashes in real-world scenarios. The simulated experiments involved 4 UAVs, with 10 runs performed for each method (with and without the proposed approach).
In contrast to the proposed approach, UAVs without the proactive distributed state estimation enhancement were unable to maintain cohesion with their teammates, leading to arbitrary drifting in different directions. This resulted in an exponentially increasing \rev{average} distance between neighboring UAVs \rev{$\overline{d}_{NB}$}.
\rev{It is calculated as the mean of all distances between UAVs considered as neighbors.}
Additionally, the drift velocity \rev{$v_{drift}$} of standalone UAVs significantly exceeded that of swarm UAVs using the proposed approach.
\rev{$v_{drift}$ is obtained as the first derivative of the UAV group's center position over the time.}
The experiment was repeated with a swarm of 8 UAVs to investigate the influence of swarm size on the final behavior. 
\rev{
  Since IMU-derived accelerations are biased, they are assigned lower trust than position measurements while still enabling rapid state updates. As shown in \autoref{fig:sim}, increasing the number of UAVs reduces velocity drift, indicating that the collective estimation framework mitigates IMU bias acting on individuals, enhancing long-term state stability.
}

\section{CONCLUSION}
This work establishes a new foundation for state estimation in UAV swarms, demonstrating that robust state awareness can be achieved using solely relative position measurements, even in the absence of environmental localization anchors such as GNSS or VIO.
By leveraging mutual information within the swarm, our approach delivers precise and consistent lateral state estimation for each UAV in the local constellation, preserving swarm cohesion in environments where external localization sources are unreliable or unavailable.
The proposed method thus enables swarms to operate autonomously and resiliently in complex and challenging environments, with robustness to disturbances affecting individual UAVs.
As validated through extensive simulations and real-world experiments, the degradation in state estimation performance is compensated by the proposed method, resulting in a uniform translational drift of the swarm as a whole.
These findings open pathways for deploying UAV swarms in a broader range of mission-critical applications, where cooperative task execution is essential but environmental localization anchors cannot be guaranteed.


\bibliographystyle{IEEEtran}
\bibliography{bibliography.bib}

\end{document}